\documentclass{article}
\usepackage{spconf,amsmath,graphicx}
\usepackage{array}
\usepackage{amsmath,amssymb}
\DeclareMathOperator{\E}{\mathbb{E}}
\usepackage{cite}
\usepackage{hyperref}
\title{GENERATIVE GUIDING BLOCK: SYNTHESIZING REALISTIC LOOKING VARIANTS CAPABLE OF EVEN LARGE CHANGE DEMANDS}
\name{Minho Park, Hak Gu Kim, and Yong Man Ro$^{*}$\thanks{* Corresponding author (ymro@ee.kaist.ac.kr). This work was partly supported by IITP grant (No. 2017-0-00780).}}
\address{Image and Video Systems Lab, School of Electrical Engineering, KAIST, South Korea}
\begin{document}
%\ninept
%
\maketitle
\begin{abstract}
Realistic image synthesis is to generate an image that is perceptually indistinguishable from an actual image. Generating realistic looking images with large variations (e.g., large spatial deformations and large pose change), however, is very challenging. Handing large variations as well as preserving appearance needs to be taken into account in the realistic looking image generation. In this paper, we propose a novel realistic looking image synthesis method, especially in large change demands. To do that, we devise generative guiding blocks. The proposed generative guiding block includes realistic appearance preserving discriminator and naturalistic variation transforming discriminator. By taking the proposed generative guiding blocks into generative model, the latent features at the layer of generative model are enhanced to synthesize both realistic looking- and target variation- image. With qualitative and quantitative evaluation in experiments, we demonstrated the effectiveness of the proposed generative guiding blocks, compared to the state-of-the-arts.  
\end{abstract}
\begin{keywords}
Deep learning, adversarial learning, variation image synthesis, and feature enhancement
\end{keywords}
\section{Introduction}
\label{sec:intro}

Generating realistic-looking images draws great attention and considered as an important task in generative models for image synthesis. Recently, deep learning-based generative models have achieved remarkable success in various synthesis tasks such as face, human, and scene generation. In data acquisition, it is time consuming and costly to collect or capture the images with desired variations (e.g., pose, illumination, facial expression, and viewpoint). Generative models that can automatically synthesize images with the desired variations are needed in practice.  

For generating realistic-looking images of objects, it is required to understand both their appearance and variants. The object has inherent appearance properties characterized by color and texture such as hair color and fashion style. On the other hand, there are variants including the shape and geometrical layout of the object. One of the most challenging points in the image generation is to preserve the appearance properties of input image (e.g., color, texture, the identity of person) while performing spatial deformation according to variants (e.g., pose variation and  illumination variation).

For this task, so far, various methods have been proposed based on Variational
Auto-Encoders (VAEs) \cite{journals/corr/KingmaW13}, Generative Adversarial Networks (GANs) \cite{NIPS2014_5423} and
Autoregressive models (ARMs) (e.g., PixelRNN \cite{VanDenOord:2016:PRN:3045390.3045575}) \cite{yeh2017semantic, NeuralFace2017, pathakCVPR16context, NIPS2016_6125, inproceedings, 10.1007/978-3-319-46484-8_16, 8633862, 8649739, 8462388}. Recently, a wide range
of methods including conditional GANs \cite{Mirza2014ConditionalGA} or conditional VAEs \cite{NIPS2015_5775} have been
proposed for synthesizing the images whose appearances depend on a given
conditioning variable (e.g., label). However, most of them could not deal with
the large variations (e.g., large spatial deformation \cite{Siarohin_2018_CVPR}) between the input and
the target image while preserving the appearance of a given input. Due to the
high dimensionality of images and the complex configuration of image contents,
it is difficult for a complete end-to-end framework to generate both the correct
target variation and the detailed appearance simultaneously \cite{ma2017pose,ma2018disentangled, Zhao:2018:MIG:3240508.3240536,Neverova_2018_ECCV}.

In this paper, we focus on realistic appearance and naturalistic variation in target image generation. The generative features are enhanced with appearance preservation and variant transformation. Our objective is to propose new generation method that addresses two problems, which are realistic appearance and naturalistic large-variation. To cope with the problems, we propose a novel generative guiding blocks (GGBs). Each generative guiding block consists of realistic appearance preserving discriminator (RAPD) and naturalistic variation transforming discriminator (NVTD). In the proposed RAPD, to preserve the object appearance of input image (e.g., identity of person), the overall image distribution is considered by determining whether the appearance is preserved in the target image or not. Simultaneously, in the proposed NVTD, to generate the target image with large variation, the change information of deformation is considered by focusing on the variation between the input and the generated target image. We hierarchically integrate the proposed GGBs with the decoding module of the generator to enhance generative feature in multiple resolution levels. The proposed generative model with GGBs enables to synthesize the realistic-looking image robustly even with large variations while maintaining naturalistic variants. Experimental results showed the effectiveness of the proposed GGBs.

\begin{figure*}[t]
\centerline{\includegraphics[width=0.9\linewidth]{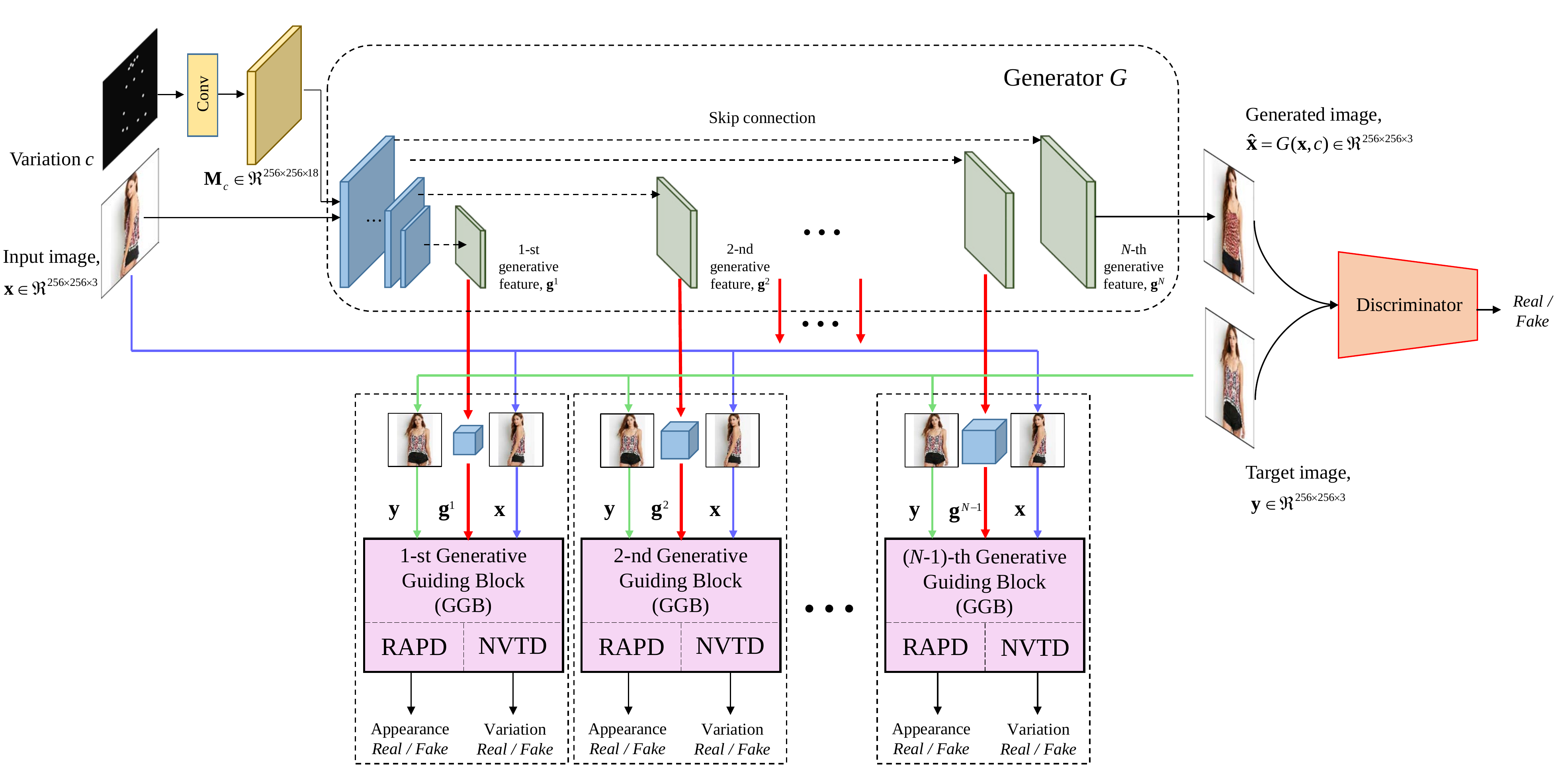}}
\vspace{-4mm}
   \caption{Overall Architecture of the proposed generative model with generative guiding blocks (GGBs). Note GBBs is hierarchically integrated in decoding modules of generator in multiple resolution levels. }
\label{fig1}
\vspace{-4.5mm}
\end{figure*}

The rest of this paper is organized as follows. In section 2, we describe the proposed generative model with GGBs. In section 3, the experimental results are presented. Finally, conclusion is drawn in section 4.

\section{PROPOSED METHOD}
\label{sec:format}

Fig. 1 shows the proposed generative model with generative guiding blocks (GGBs). The generator synthesizes the fake image having the appearance of the input image and the target variants. The discriminator determines whether the fake image is real or not. As shown in Fig. 1, the generative guiding blocks (GGBs) are attached to multi-level generative features of multiple layers in the decoder of generator. The GGBs determine whether the generated multi-resolution images have realistic appearance (operated by RAPD in GGB) and naturalistic variation (operated by NVTD in GGB). Variant transformation is performed hierarchically in a multi-resolution manner so that the proposed generator can process large variant demand. In the following subsections, we describe in detail about the generator, discriminator and GGBs.

\subsection{Generative model with discriminator}
\label{ssec:subhead}

Let ${\mathbf{x}}\in {\rm I\!R}^{256 \times 256 \times 3}$ denote the input image and ${\mathbf{y}}\in {\rm I\!R}^{256 \times 256 \times 3}$ denote the ground-truth target image. $c$ denotes the target variation and $\hat{\mathbf{x}}\in{\rm I\!R}^{256 \times 256 \times 3}$ (i.e. $G(\mathbf{x},{c})$) denotes the generated image. Let $\mathbf{g}^{n}$ denote $n$-th generative feature. Let $G$ denote the generator, $D$ denote the discriminator and ${\mathbf{M}_{c}}\in {\rm I\!R}^{256 \times 256 \times 3}$ denote the label map which is encoded from $c$. By encoding $c$, abundant condition information of the desired variation is provided to the $G$. In this paper, a U-Net-like structure is employed as $G$ \cite{RFB15a, 10.1007/978-3-030-05716-9_1}. The encoder and decoder of $G$ consist of 7 convolution layers and deconvolution layers, respectively (i.e. $N$=7) with $4\times4$ kernel and stride of 2. $D$ consists of 5 convolution layers with $4\times4$ kernel and stride of 2.

With an adversarial learning \cite{NIPS2014_5423}, $D$ determines whether the $\hat{\mathbf{x}}$̂ is a realistic-looking or not, comparing with ${\mathbf{y}}$. The objective functions of $D$ can be written as 
\begin{equation}\label{eq1}
\begin{aligned}
    \mathcal{L}_{D}=&-\E_{\mathbf{y}\sim p_{\mathbf{y}}}[\text{log}(D(\mathbf{y}))]\\ &-\E_{\mathbf{x}\sim p_{\mathbf{x}}}[\text{log}(1-D(G(\mathbf{x}, c)))].
\end{aligned}
\end{equation}

On the other hand, $G$ tries to fool $D$ by generating the realistic image. To that end, the loss of the generator is composed of two terms, which are the realism loss, $\ell_{real}$, and the reconstruction loss, $\ell_{rec}$. The realism loss can be written as
\begin{equation}\label{eq2}
    \mathcal{} \ell_{real}=-\E_{\mathbf{x} \sim p_{\mathbf{x}}}[\text{log}(D(G(\mathbf{x},c)))]
\end{equation}

The reconstruction loss between the ground-truth target image and the generated image at $n$-th level, $\ell^{n}_{rec}$ , in the decoder can be written as
\begin{equation}\label{eq3}
    \mathcal{} \ell^{n}_{rec}=\E_{\mathbf{x} \sim p_{\mathbf{x}}}[\Vert \mathbf{y}^{n}-\hat{\mathbf{x}}^{n} \Vert_{1}],
\end{equation}
where $\hat{\mathbf{x}}^{n}$  indicates a generated image from ${\mathbf{g}^{n}}$ and ${\mathbf{y}^{n}}$ indicates an image downsized to the same resolution of $\hat{\mathbf{x}}^{n}$ from ${\mathbf{y}}$ (as shown in Fig. 2).

Finally, the total loss function of the proposed generator, $G$, can be defined as a combination of the realism loss and the reconstruction loss.
\begin{equation}\label{eq4}
    \mathcal{L}_{G} = \lambda_{real} \ell_{real}+\ell^{N}_{rec},
\end{equation}
where $\lambda_{real}$ is a weight parameter to control the balance between $\ell_{real}$ and  $\ell^{N}_{rec}$.

\begin{figure}[t]
\centerline{\includegraphics[width=0.95\linewidth]{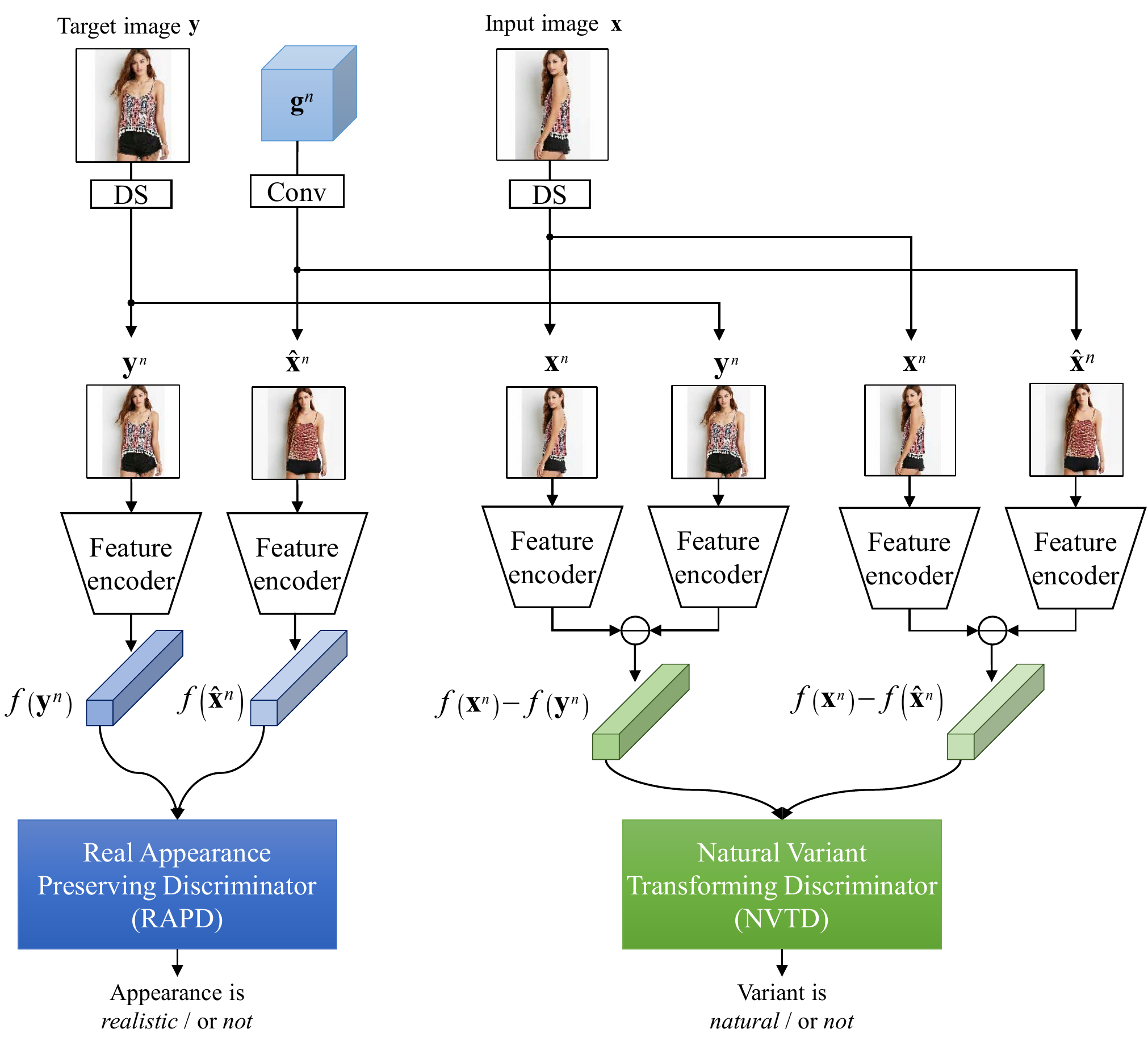}}
\vspace{-3mm}
   \caption{The architecture of the proposed $n$-th GGB.}
  \vspace{-4.5mm}
\label{fig2}
\end{figure}

\subsection{Generative Guiding Block for realistic appearance and naturalistic variation}
\label{ssec:subhead}

Fig. 2 shows the architecture of the proposed $n$-th GGB, which consists of a realistic appearance preserving discriminator (RAPD), ${D}_{RAPD}$, and a naturalistic variation transforming discriminator (NVTD), ${D}_{NVTD}$. The GGBs are attached on the multi-level generative features of multiple layers in the decoder as shown in Fig. 1. Let ${\mathbf{x}^{n}}$ denote an image downsized to the same resolution of $\hat{\mathbf{x}}^{n}$ from ${\mathbf{x}}$. Let $f(\cdot)$ denote the feature encoder. In this paper, ${D_{RAPD}}$ and ${D_{NVTD}}$ consist of 3 convolution layers. The feature encoder consists of 2 convolution layers with 4$\times$4 kernel and stride of 2.  

First, to deal with feature information of ${\mathbf{x}^{n}}$, $\hat{\mathbf{x}}^{n}$ and ${\mathbf{y}^{n}}$, the images are encoded to the latent feature, ${f}({\mathbf{x}^{n}})$, ${f}(\hat{\mathbf{x}}^{n})$ and ${f}({\mathbf{y}^{n}})$. After that, ${D}_{RAPD}$ distinguishes whether the encoded features, ${f}(\hat{\mathbf{x}}^{n})$ and ${f}({\mathbf{y}^{n}})$, are realistic or not. As shown in Fig. 2, ${D}_{NVTD}$ distinguishes whether the residual information of encoded features (i.e., $\mathbf{d}_{real}^{n} = {f}({\mathbf{x}^{n}}) - {f}({\mathbf{y}^{n}})$ and $\mathbf{d}_{fake}^{n} = {f}({\mathbf{x}^{n}}) - {f}(\hat{\mathbf{x}}^{n})$) is realistic or not. The reason that the input of ${D}_{NVTD}$ is residual information is to make ${D}_{NVTD}$ focus on only the target variation. $G$ tries to fool ${D}_{RAPD}$, so that $\hat{\mathbf{x}}^{n}$ mimics the data distribution of ${\mathbf{y}^{n}}$. Through this process, ${\mathbf{g}^{n}}$ is enhanced for generating appearance realistic image. Also, $G$ tries to fool ${D}_{NVTD}$, so that $\mathbf{d}_{fake}^{n}$ tries to follow $\mathbf{d}_{real}^{n}$. ${\mathbf{g}^{n}}$ is enhanced for generating the image with naturalistic variation as well.

The discriminators in GGB, ${D}_{RAPD}$ and ${D}_{NVTD}$, are trained by adversarial learning with $G$. Therefore, we adopt generative adversarial loss. First, the objective function of ${D}_{RAPD}$ is defined as
\begin{equation}\label{eq5}
\begin{aligned}
    \mathcal{L}_{D_{RAPD}}^{n} =&-\E_{\mathbf{y}\sim p_{\mathbf{y}}}[\text{log}({D_{RAPD}^{n}}(f(\mathbf{y}^{n})))]\\ &-\E_{\mathbf{x}\sim p_{\mathbf{x}}}[\text{log}(1-{D_{RAPD}^{n}}(f(\hat{\mathbf{x}}^{n})))],
\end{aligned}
\end{equation}
where ${D_{RAPD}^{n}}$ indicates $D_{RAPD}$ in $n$-th GGB.
Similarly, the objective function of $D_{NVTD}$ is defined as 
\begin{equation}\label{eq6}
\begin{aligned}
	\mathcal{L}_{D_{NVTD}}^{n} &=-\E_{\mathbf{x}\sim p_{\mathbf{x}}, \mathbf{y} \sim p_{\mathbf{y}}}[\text{log}({D_{NVTD}^{n}}({\mathbf{d}_{real}^{n}}))]\\ &-\E_{\mathbf{x}\sim p_{\mathbf{x}}}[\text{log}(1-{D_{NVTD}^{n}}({\mathbf{d}_{fake}^{n}}))],
\end{aligned}
\end{equation}
where ${D_{NVTD}^{n}}$ indicates $D_{NVTD}$ in $n$-th GGB. 

${D_{RAPD}^{n}}$ and ${D_{NVTD}^{n}}$ are trained to minimize ${\mathcal{L}_{D_{RAPD}}^{n}}$ and ${\mathcal{L}_{D_{NVTD}}^{n}}$, respectively. Contrary, $G$ with GGBs is trained to minimize ${\ell_{RAPD}^{n}}$  and  ${\ell_{NVTD}^{n}}$ for learning to fool ${D_{RAPD}^{n}}$ and ${D_{NVTD}^{n}}$. These objective functions can be written as
\begin{equation}\label{eq7}
    \mathcal{} {\ell_{RAPD}^{n}}=-\E_{\mathbf{x} \sim p_{\mathbf{x}}}[\text{log}({D_{RAPD}^{n}}(f(\hat{\mathbf{x}}^{n})))],
\end{equation}
\begin{equation}\label{eq8}
    \mathcal{} {\ell_{NVTD}^{n}}=-\E_{\mathbf{x} \sim p_{\mathbf{x}}}[\text{log}({D_{NVTD}^{n}}({\mathbf{d}_{fake}^{n}}))].
\end{equation}

In particular, to preserve the appearance information, we adopt the L1 norm as our reconstruction loss, Eq. \ref{eq3}. Finally, the objective function of G with our GGBs is defined as
\begin{equation}\label{eq9}
\mathcal{L}_{GGB}=\sum^{N-1}_{n=1}{\lambda_{RAPD}^{n}}{\ell_{RAPD}^{n}}+{\lambda_{NVTD}^{n}}{\ell_{NVTD}^{n}}+{\ell_{rec}^{n}},
\end{equation}
where $\Sigma$ is used for weighted sum of multi-level GGB losses. 

\begin{table}[t!]
\vspace{-5mm}
\renewcommand{\arraystretch}{1.1}
\caption{Quantitative comparison with the state-of-the-art methods on DeepFashion dataset.}
\label{my-label}
\centering
\small
\begin{tabular}{c p{2.5cm} p{1.5cm} c}
\hline\hline
\centering \textbf{Model}  & \centering \textbf{SSIM}  &\centering  \textbf{IS} & \\ \hline\hline
Disentangled\cite{ma2018disentangled} &\centering 0.614 & \centering3.23 &\\
VariGAN\cite{Zhao:2018:MIG:3240508.3240536}      &\centering 0.620 & \centering 3.03 &\\
PG$^{2}$\cite{ma2017pose}     &\centering 0.762 &\centering 3.09 &\\
DPT\cite{Neverova_2018_ECCV}          &\centering 0.769 &\centering 3.17 &\\ \hline
\textbf{Ours}                  &\centering \textbf{0.799} &\centering \textbf{3.26} &\\ \hline
\end{tabular}%
\end{table}

\begin{table}[t!]
\renewcommand{\arraystretch}{1.1}
\vspace{-5mm}
\caption{Effectiveness of using both RAPD/NVTD and multiple GGBs}
\label{my-label}
\centering
\small
\begin{tabular}{c p{2.5cm} p{1.5cm} c}
\hline\hline
\centering \textbf{Model}  & \centering \textbf{SSIM}  &\centering  \textbf{IS} & \\ \hline\hline
Ours w/o GGBs         &\centering 0.705 &\centering 2.81 &\\
Ours w/o RAPD         &\centering 0.709 &\centering 2.72 &\\
Ours w/o NVTD         &\centering 0.714 &\centering 2.73 &\\ \hline
Ours with 1 GGB       &\centering 0.780 &\centering 3.14 &\\
Ours with 2 GGBs      &\centering 0.793 &\centering 3.15 &\\ \hline
\textbf{Ours}         &\centering \textbf{0.799} &\centering \textbf{3.26} &\\ \hline
\end{tabular}%
\vspace{-3mm}
\end{table}

\begin{figure*}[t]
\centerline{\includegraphics[width=0.55\linewidth]{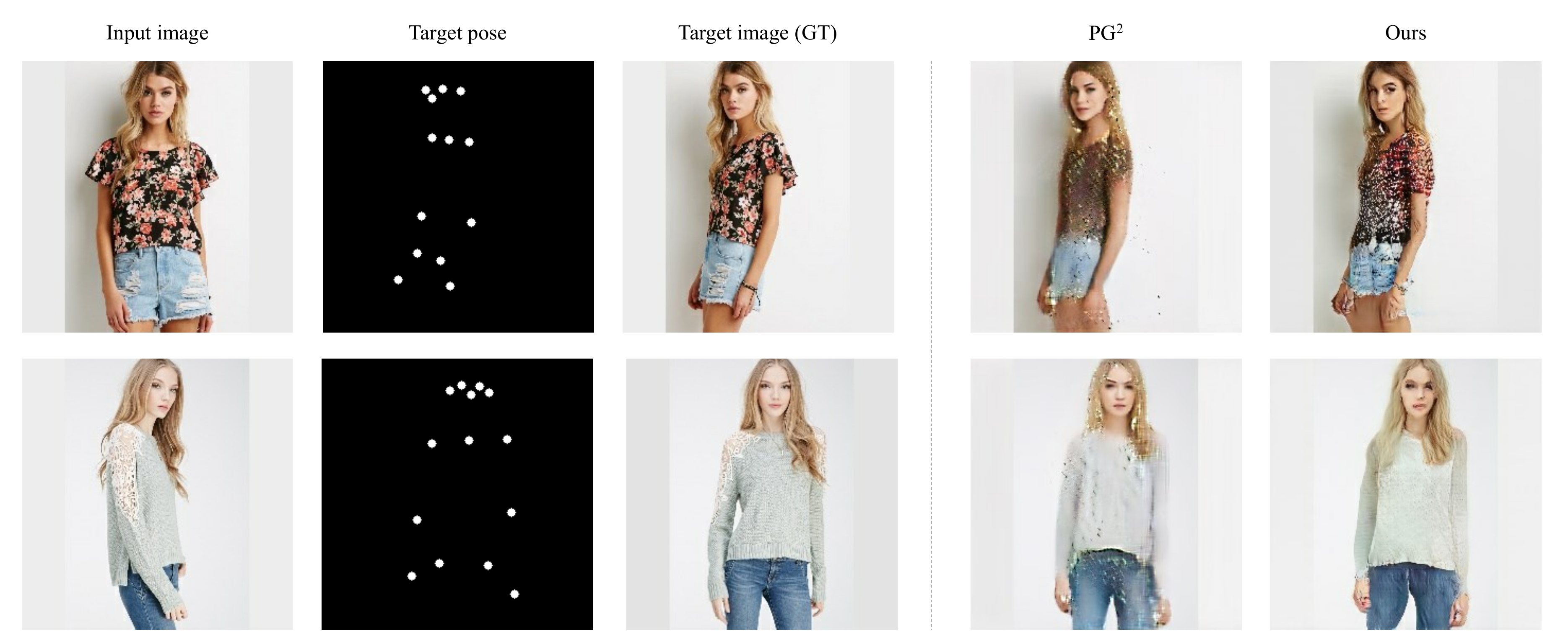}}
    \vspace{-4mm}
   \caption{Qualitative comparison on DeepFashion dataset between the results obtained by our approach and PG$^{2}$\cite{ma2017pose}.}
\label{fig3}
\vspace{-4mm}
\end{figure*}

\begin{figure*}[t]
\centerline{\includegraphics[width=0.65\linewidth]{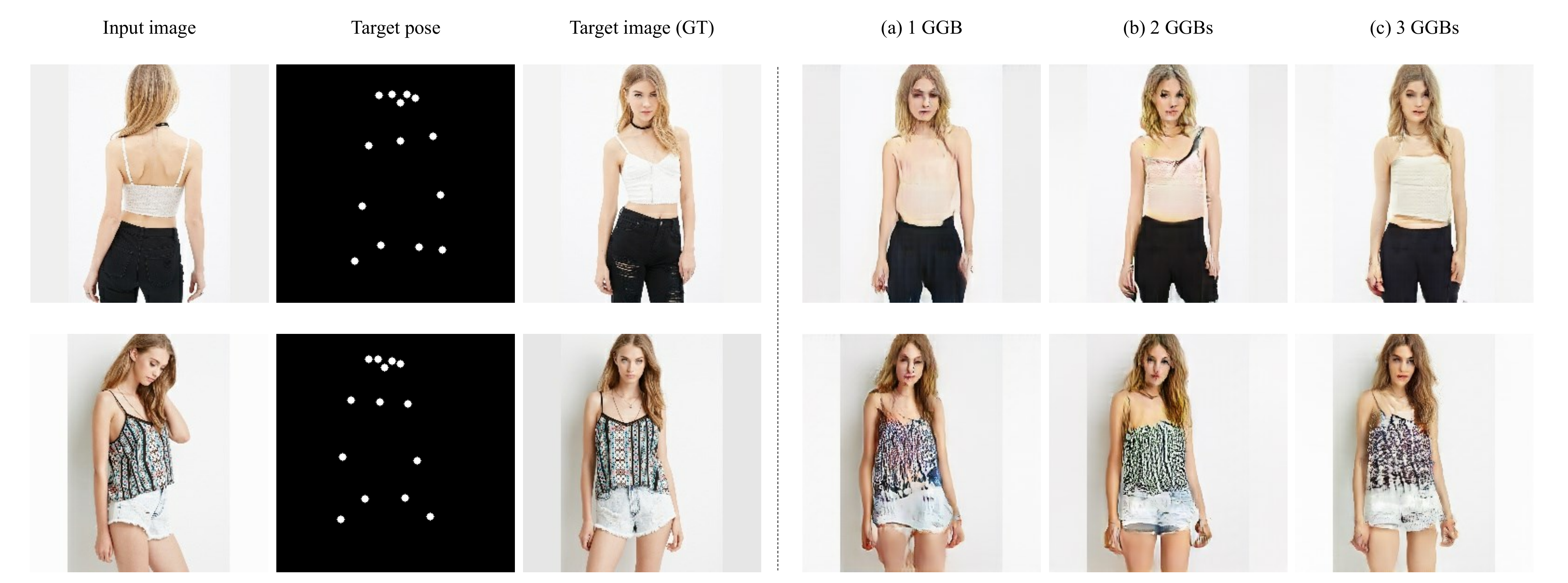}}
   \vspace{-4mm}
   \caption{Generated human pose images obtained by our model on DeepFashion dataset when it is trained with (a) only 6-th GGB, (b) 5-th and 6-th GGBs, (c) 4-th, 5-th and 6-th GGBs.}
\label{fig4}
\vspace{-4.5mm}
\end{figure*}

\vspace{-1.5mm}
\subsection{Training strategy}
\label{ssec:subhead}

Every iteration, $\mathbf{x}$ and $c$ are given to $G$. Then, $G$ generates $\hat{\mathbf{x}}$̂. In the $D$, $\mathcal{L}_{D}$ is calculated with $\hat{\mathbf{x}}$̂ and ${\mathbf{y}}$ (see Eq.\ref{eq1}). In the $n$-th GGB, ${\mathcal{L}_{D_{RAPD}}^{n}}$ and ${\mathcal{L}_{D_{NVTD}}^{n}}$ are calculated with $\mathbf{x}^{n}$, $\mathbf{y}^{n}$ and $\hat{\mathbf{x}}^{n}$ (see Eq.\ref{eq5} and \ref{eq6}). After that, the weights of $D$ are updated to minimize $\mathcal{L}_{D}$. Also, the weights of $n$-th GGB are updated to minimize ${\mathcal{L}_{D_{RAPD}}^{n}}$ and ${\mathcal{L}_{D_{NVTD}}^{n}}$ ($n$=1,2,...,$N$-1). The weights of $G$ except for $\mathbf{g}^{N}$ are firstly updated to minimize $\mathcal{L}_{GGB}$ (see Eq. \ref{eq9}). Finally, the  weights of $G$ are updated to minimize $\mathcal{L}_{G}$ (see Eq. \ref{eq4}). Until the weights are optimized, this process is repeated.

\vspace{-1.5mm}
\section{EXPERIMENTS AND RESULTS}
\label{sec:pagestyle}

\vspace{-1.5mm}
\subsection{Datasets}
\label{ssec:subhead}

For verifying the effectiveness of the proposed generative model with GGBs, we used public datasets: DeepFashion \cite{conf/cvpr/LiuLQWT16}. This dataset consists of 52,712 in-shop clothes images with 256$\times$256 resolution. As similar to \cite{ma2017pose}, for the training set, we have 146,680 pairs. Each pair is composed of two images of the same identity but different poses. For the test set, we randomly selected 12,800 pairs from the test set. To use the human pose landmark of DeepFashion data as the target variation, we applied a state-of-the-art pose estimation \cite{8099626}, as in \cite{ma2017pose}.

\vspace{-1.7mm}
\subsection{Implementation details}
\label{ssec:subhead}

We used Adam optimizer \cite{journals/corr/KingmaB14} with ${\beta}_{1}$ = 0.5, ${\beta}_{2}$ = 0.999, the batch size of 8, and learning rate of 0.0002 to train proposed models. In our experiment, we attached three GGBs on the generative features with 32 $\times$ 32, 64 $\times$ 64 and 128 $\times$ 128 resolutions (i.e. $\mathbf{g}^{4}$, $\mathbf{g}^{5}$ and $\mathbf{g}^{6}$). We empirically set ${\lambda_{real}}$ = 0.02 and ${\lambda_{RAPD}^{n}}$ = ${\lambda_{NVTD}^{n}}$ = 0.01.  

\vspace{-1.7mm}
\subsection{Performance evaluation}
\label{ssec:subhead}

Fig. 3 shows comparison between generated images by our model and those by the state-of-the-art model, PG$^{2}$\cite{ma2017pose}. To obtain the results of PG$^{2}$, we used pretrained weight provided by the author of PG$^{2}$. As shown in Fig. 3, in the results of PG$^{2}$, hair and clothes were blurred a lot. Thus the appearance information was not preserved well. On the other hand, the appearances were preserved well in ours. 
Fig. 4 shows the effectiveness of refining multi-level features using GGBs. '1 GGB' indicates the generative model with only 6-th GGB. '2 GGBs' indicates the generative model with 5-th and 6-th GGBs. '3 GGBs' indicates the generative model with 4-th, 5-th and 6-th GGBs, same as proposed model. The more GGBs were used in generative model training, the clearer the images and the better the appearance preserved.
Table 1 and 2 show the quantitative results of state-of-the-art models \cite{ma2017pose, ma2018disentangled, Zhao:2018:MIG:3240508.3240536,Neverova_2018_ECCV} and the proposed model by measuring Structural Similarity (SSIM) \cite{journals/tip/WangBSS04} and Inception scores (IS) \cite{NIPS2016_6125}. As seen in Table 1, the proposed method outperformed the state-of-the-art method. In table 2, 'w/o GGBs' indicates training generative model without any GGB. 'w/o RAPD' and 'w/o NVTD' indicate that there are only NVTD and RAPD in GGB, respectively. As seen in Table 2, the proposed model (i.e. 3 GGBs are used, RAPD and NVTD in GGB) provided the highest performance.

\vspace{-2.8mm}
\section{CONCLUSION}
\vspace{-2mm}
\label{sec:typestyle}

In this paper, we proposed a novel Generative Guiding Block for synthesizing realistic looking images with the
large variations while preserving the appearance properties.
The proposed GGB consisted of two critic networks which
were RAPD for maintaining the appearance characteristic and
NVTD for applying the target variants. By hierarchically integrating the proposed GGBs with the generator, the proposed
GGBs could enhance the generative features in the decoder
from coarse to fine. The experimental results showed that the
proposed method outperformed the state-of-the-art methods.
Also, the effectiveness of components of GGB (i.e. RAPD
and NVTD) and hierarchical multi-level features were shown.

\bibliographystyle{IEEEbib}

\end{document}